\pdfoutput=1

\documentclass[11pt]{article}

\usepackage[final]{acl}

\usepackage{times}
\usepackage{latexsym}

\usepackage[T1]{fontenc}

\usepackage[utf8]{inputenc}

\usepackage{microtype}

\usepackage{inconsolata}

\usepackage{graphicx}

\usepackage{times}
\usepackage{latexsym}
\usepackage{times}
\usepackage{latexsym}
\usepackage{booktabs}
\usepackage{boxedminipage}
\usepackage{amssymb}
\usepackage{amsmath}
\usepackage{todonotes} 
\usepackage{graphicx}

\usepackage{hyperref}
\usepackage{inconsolata}
\usepackage{pifont}

\usepackage{tabu}
\usepackage{fontawesome}
\usepackage{amsmath}
\usepackage{amsfonts}
\usepackage{amssymb}
\usepackage{enumitem}
\usepackage{listings}
\usepackage{xstring}
\usepackage{graphicx}
\usepackage{pbox}
\usepackage{subcaption}
\usepackage{epstopdf}
\usepackage{xstring}
\usepackage{multirow}

\usepackage{soul}
\usepackage{color}
\usepackage{graphicx}
\usepackage{multirow}
\usepackage{comment}
\usepackage{listings} 
\usepackage{graphicx}
\usepackage{subcaption}
\usepackage{verbatim}

\usepackage{amssymb}

\usepackage{xcolor} 
\definecolor{lightblue}{rgb}{.50,.90,0.51}
\definecolor{tri}{rgb}{.25,.88,.82}
\definecolor{lilac}{rgb}{0.85,0.64,0.85}
\definecolor{atomictangerine}{rgb}{1.0, 0.6, 0.4}

\title{\textsf{\textbf{PropXplain}}: Can LLMs Enable Explainable Propaganda Detection?}

\author{Maram Hasanain$^1$\textsuperscript{$\dagger$}, Md Arid Hasan$^1$\thanks{~The contribution was made while the author was a contributor at the Qatar Computing Research Institute.}, Mohamed Bayan Kmainasi$^1$$^*$, Elisa Sartori$^2$, \\
{\bf Ali Ezzat Shahroor$^1$$^*$, Giovanni Da San Martino$^2$, Firoj Alam$^1$\thanks{~Corresponding authors.}}\\
  $^1$Qatar Computing Research Institute, Qatar, 
  $^2$University of Padova, Italy \\
  {\tt fialam@hbku.edu.qa, giovanni.dasanmartino@unipd.it}
\\}

\begin{document}
\maketitle
\begin{abstract}
There has been significant research on propagandistic content detection across different modalities and languages. However, most studies have primarily focused on detection, with little attention given to explanations justifying the predicted label. This is largely due to the lack of resources that provide explanations alongside annotated labels. To address this issue, we propose a multilingual (i.e., Arabic and English) explanation-enhanced dataset, the \textit{first} of its kind. Additionally, we introduce an explanation-enhanced LLM for both label detection and rationale-based explanation generation. Our findings indicate that the model performs comparably while also generating explanations. We will make the dataset and experimental resources publicly available for the research community.\footnote{
\url{https://github.com/firojalam/PropXplain}
}


\end{abstract}

\section{Introduction}
\label{sec:introduction}

The proliferation of propagandistic content in online and social media poses a significant challenge to information credibility, shaping public opinion through manipulative rhetorical strategies~\cite{da-san-martino-etal-2019-fine}. Automatic propaganda detection has been an active area of research, with studies focusing on textual~\cite{barron2019proppy}, multimodal~\cite{ACL2021:propaganda:memes}, and multilingual approaches~\cite{piskorski-etal-2023-multilingual,zhang2022cross}. However, most existing systems lack rational explanations that could improve media literacy and calibrate trust in predictions.
 
\begin{figure}[t]
    \centering
    \includegraphics[scale=0.25]{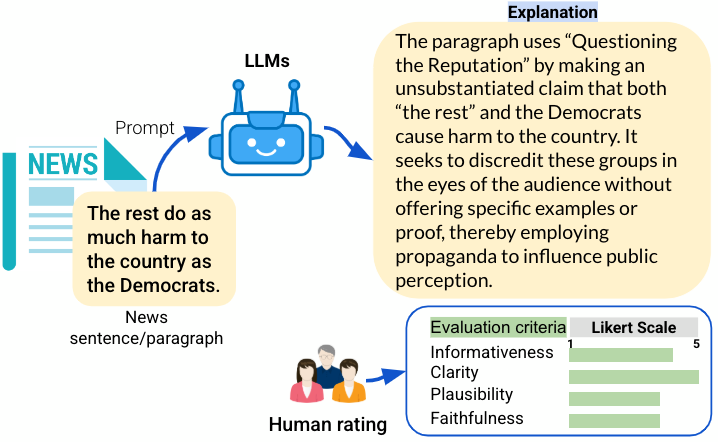}
    \vspace{-0.3cm}
    \caption{Example of a news sentence and its explanation and quality assessment process.}
    \label{fig:expl_example}
    \vspace{-0.7cm}
\end{figure}

~\citet{RANLP2021:propaganda:interpretable} developed interpretable models for propaganda detection in news articles, combining qualitative features with pre-trained language models to enhance transparency. More recently,~\citet{10.1145/3613904.3642805} conducted a user study in which GPT-4 was used for propaganda detection and explanation generation. They demonstrate that explanations foster critical thinking and highlight their importance. 
However, the current literature has paid little to no attention to developing datasets that include explanations alongside annotated propaganda labels. To address this gap, we propose a large multilingual (i.e., Arabic and English) explanation-enhanced dataset for propaganda detection. We build upon existing datasets, including ArPro~\cite{hasanain2024can} and the SemEval-2023 English dataset~\cite{piskorski-etal-2023-semeval}, enhancing them with explanations. Given the complexity of manually generating explanations and the higher reliability reported for GPT-4-based explanation generation~\cite{wang_evaluating_2023}, we opted to use a stronger LLM for explanation generation and manually checked for quality assurance. Figure \ref{fig:expl_example} shows an example of a news sentence, its explanation, and human evaluation process. The developed dataset can be used to train specialized LLMs for propaganda detection and to provide explanations for their predictions.  
To this end, our contributions to this study are as follows:
\begin{itemize}[noitemsep,topsep=0pt,labelsep=.5em]
    \item We introduce an explanation-enhanced dataset for propaganda detection, consisting of approximately 21k and 6k news paragraphs and tweets for Arabic and English, respectively.  
    \item To ensure the quality of the LLM-generated explanations, we manually evaluate  explanations of the test set for each language.  
    \item Our comparative experiments show that the proposed LLM matches transformer-based models in performance while additionally providing explanations for its predictions.  
\end{itemize}

\section{Dataset}
\label{sec:dataset}

We investigate LLMs' ability for explainable propaganda detection in both a high-resource language (English) and a lower-resource language (Arabic). In this work, we extend existing datasets with natural language annotation explanations generated by OpenAI o1, and evaluated by humans. 

\begin{table}[t]
\centering
\resizebox{\columnwidth}{!}{%
\begin{tabular}{lrrrrr}
\toprule
\textbf{Split} & \multicolumn{1}{c}{\textbf{\# Articles}} & \multicolumn{1}{c}{\textbf{\#items}} & \multicolumn{1}{c}{\textbf{Avg (W)}} & \multicolumn{1}{c}{\textbf{Avg Exp. (W)}} & \multicolumn{1}{c}{\textbf{\% Prop.}} \\
\hline
\multicolumn{6}{c}{\textbf{Arabic}}\\
\hline
Train & 8,103 &18,453    & 32.4     & 48.1 &63.8\%  \\
Dev   & 822   &1,318     & 32.6     & 47.9 &64.4\%  \\
Test  & 835   &1,326     & 35.1     & 48.7 &61.3\%  \\ \midrule
\textbf{Total} & \textbf{8,913$^{*}$} & \textbf{21,097}  & \textbf{32.6}     & 48.1 &\textbf{63.7\%}\\
\hline
\multicolumn{6}{c}{\textbf{English}}\\
\hline
Train & 250   &4,472     & 24.0     & 61.2 &26.9\%  \\
Dev   & 204   &621       & 23.9     & 61.6 &27.9\%  \\
Test  & 225   &922       & 23.7     & 61.2 &27.9\%  \\ \midrule
\textbf{Total} & \textbf{250$^{*}$} & \textbf{6,015}   & \textbf{24.0}     & \textbf{61.2} &\textbf{27.2\%}\\
\bottomrule
\end{tabular}%
}
\vspace{-0.3cm}
\caption{Distribution of Arabic and English datasets. Exp.: explanation. Data items: annotated data elements including paragraphs and tweets. $^{*}$ Total unique articles. Prop.: Propagandistic. W: \# Words}
\label{tab:data_stats}
\vspace{-0.4cm}
\end{table}

\subsection{Arabic Propaganda Dataset}
Building upon the ArPro Arabic dataset~\cite{hasanain2024can}, we follow the same annotation approach to build a larger dataset by collecting and annotating 7$K$ paragraphs. 
 Furthermore, this extension includes collecting and annotating tweets, to examine propaganda use in social media. Eventually, our Arabic dataset comprises two types of annotated documents: tweets and news paragraphs. The news paragraphs are extracted from articles published by 300 distinct news agencies, capturing a broad spectrum of Arabic news sources. It covers a diverse range of writing styles and topics including 14 different topics such as  news,  politics, human rights, and science and technology.
As for the tweets subset, we start from a manually constructed set of 14 keywords and phrases, covering the topic of Israeli-Palestinian war, targeting sub-topics popular during October and early November 2023. We use Twitter's search API to search for tweets posted during the second week of November 2023 and matching the collected phrases, 
resulting in 5.7$K$ tweets to annotate. 

Data was annotated following a two-phase approach~\cite{hasanain2024can}. In the first phase, $3$ annotators independently examine each data item (paragraph or tweet) and label it with
propagandistic techniques. In the second phase, $2$ expert annotators examine annotations from the first phase and resolve any conflicts. Finally, the dataset set was split into training, development, and testing subsets following a stratified sampling approach.

\subsection{English Propaganda Dataset}

The English dataset is composed of $250$ articles, collected from $42$ unique news sources, coming from all political positions. The articles are manually cleaned of any artifacts that are incorrectly included during collection, such as links. The articles include topics that trended in the late 2023 and early 2024, with discussions of politics and the Israeli-Palestinian war covering $60\%$ of the articles.
Each article is annotated by at least $2$ annotators and reviewed by $1$ curator, whose task is to resolve inconsistencies between annotations. During the whole process, random checks of the annotations are carried out to verify the quality and give feedback on inaccuracies. To create the dataset, the articles are divided into sentences and split into three subsets: training, development and testing.

Note that these datasets are annotated for fine-grained propaganda detection; however, for this study, we perform classification and explanation generation in a binary setup.

\subsection{Explanation Generation}
We use OpenAI o1 to generate natural language explanations for gold propaganda annotations. This LLM is designed to have superior reasoning capabilities\footnote{\url{https://openai.com/index/introducing-openai-o1-preview/}} which we believe are required for the task at hand. During pilot studies, we experimented with another highly-effective LLM, GPT-4o and a variety of prompts. Our manual evaluation of different samples in English and Arabic 
revealed that explanations generated by OpenAI o1 are better on average (following the quality assessment described in the next section). Eventually, the following prompt is used for explanation generation: \textit{``Generate one complete explanation shorter than 100 words on why the paragraph as a whole is [gold label (propagandistic/not propagandistic)]. Be very specific in this full explanation to the paragraph at hand. Your explanation must be fully in [language].''}
Note that we used the original coarse-grained technique labels when generating gold explanations. However, for the current study, we mapped these labels to binary categories to evaluate the impact of explanations on classification and interpretability. For propagandistic texts, we generated explanations for both the individual spans with technique labels and the entire text, incorporating the relevant techniques. For non-propagandistic texts, we provided explanations justifying the label (not-propagandistic). 

\paragraph{Quality of Generated Explanations}
We verify the quality of the generated explanations by human evaluation. We used a 5-point Likert scale for various evaluation metrics selected from relevant studies on natural language explanation evaluation~\cite{huang-etal-2024-chatgpt,10.1145/3543873.3587368,10.1145/3613904.3642805}, including \textit{informativeness, clarity, plausibility, and faithfulness}. 
Evaluation was carried out for Arabic and English datasets on the full test set.
We provided detailed annotation instructions guidelines (see in Appendix \ref{sec:app_annotation_guideline}) for the human evaluators and each explanation assessed by three evaluators (see in Appendix \ref{sec:app_annotation_setup}).

In Table \ref{tab:likert_score}, we report the average scores for all evaluation metrics. We first compute the average across annotators for each explanation and then across all explanations. In addition, we also computed the annotation agreement on ordinal scales by adopting the agreement index $r^*_{wg(j)}$ \cite{james1984estimating}, which compares the observed variance in ratings to the maximum possible variance under complete disagreement (see further details in Appendix \ref{sec:app_annotation_agreement}). As presented in Table \ref{tab:likert_score_annotation_agr}, the values above 0.89 for Arabic and 0.94 for English suggest a strong agreement~\cite{o2017overview}. The results also suggest that OpenAI o1 generally generates explanations that are of high quality, considering the metrics at hand (e.g., clarity).




\begin{table}[h]
\centering
\setlength{\tabcolsep}{1pt} 
\scalebox{0.85}{%
\begin{tabular}{@{}lcccc@{}}
\toprule
\multicolumn{1}{l}{\textbf{Data}} & \multicolumn{1}{c}{\textbf{Faithfulness}} & \multicolumn{1}{c}{\textbf{Clarity}} & \multicolumn{1}{c}{\textbf{Plausibility}} & \multicolumn{1}{c}{\textbf{Informative}} \\ \midrule
Arabic & 4.35 & 4.49 & 4.42 & 4.26 \\
English & 4.72 & 4.76 & 4.71 & 4.71 \\ \bottomrule
\end{tabular}
}
\vspace{-0.1cm}
\caption{Average Likert scale value for each human evaluation metric across different sets of explanations.}
\label{tab:likert_score}
\vspace{-0.3cm}
\end{table}

\begin{table}[]
\centering
\setlength{\tabcolsep}{1pt} 
\scalebox{0.78}{%
\begin{tabular}{@{}lrrrr@{}}
\toprule
\multicolumn{1}{c}{\textbf{Dataset}} & \multicolumn{1}{c}{\textbf{Faithfulness}} & \multicolumn{1}{c}{\textbf{Clarity}} & \multicolumn{1}{c}{\textbf{Plausibility}} & \multicolumn{1}{c}{\textbf{Informative}} \\ \midrule
Arabic & 0.90 & 0.92 & 0.89 & 0.89 \\
English & 0.94 & 0.95 & 0.94 & 0.95 \\ \bottomrule
\end{tabular}
}
\vspace{-0.2cm}
\caption{Annotation agreement for each human evaluation (annotation) metric across datasets, computed using  $r^*_{wg(j)}$ index.}
\label{tab:likert_score_annotation_agr}
\vspace{-0.4cm}
\end{table}

\begin{table*}[htb!]
\resizebox{\textwidth}{!}{%
\begin{tabular}{p{4.3cm}|p{4.5cm}|p{4.5cm}|p{4.5cm}}
\toprule
\textbf{Paragraph} & \textbf{Gold} &\textbf{Llama-Base}  &\textbf{Llama-FT}\\
\hline
Every single person when I came to them saying, ‘Can I get a half billion dollar bond?’ & The paragraph is \textit{not propagandistic} because it merely conveys an individual's experience without attempting to manipulate the audience's perception or promote a specific agenda. & The text appears to be \textit{propagandistic} because it presents a personal anecdote that is likely intended to persuade or influence the audience, rather than providing a neutral statement. & The paragraph is \textit{not propagandistic} because it presents a straightforward statement without employing manipulative language or techniques.\\
\bottomrule
\end{tabular}%
}
\vspace{-0.2cm}
\caption{Generated explanations by different models.}
\label{tab:exp_example}
\vspace{-0.3cm}
\end{table*}

\begin{table}[h]
\centering
\setlength{\tabcolsep}{2pt} 
\scalebox{0.7}{
\begin{tabular}{@{}lrrr@{}}
\toprule
\textbf{Model} & \multicolumn{1}{c}{\textbf{F1$_{Micro}$}} & \multicolumn{1}{c}{\textbf{F1$_{Macro}$}} & \multicolumn{1}{l}{\textbf{F1$_{BERT}$}} \\ \midrule
\multicolumn{4}{c}{\textbf{Arabic}} \\ \midrule
AraBERT & 0.762 & 0.749 & -- \\
GPT-4o & 0.575 & 0.567 & -- \\
Llama 3.1 8B (label only)	& 0.794	& 0.780 &  \\
Llama 3.1 8B (Base) & 0.588 & 0.588 & 0.507 \\
Llama 3.1 8B (FT) & 0.775 & 0.760 & 0.706 \\ \midrule
\multicolumn{4}{c}{\textbf{English}} \\ \midrule
BERT-base & 0.772 & 0.691 & -- \\
GPT-4o & 0.649 & 0.630 & -- \\
Llama 3.1 8B (label only) &	0.766	& 0.686	& \\
Llama 3.1 8B (Base) & 0.572 & 0.562 & 0.596 \\
Llama 3.1 8B (FT) & 0.781 & 0.675 & 0.751 \\
\bottomrule
\end{tabular}
}
\vspace{-0.2cm}
\caption{Performance of the proposed model and baselines. F1$_{BERT}$ is the F1 score computed using BERTScore for the explanation.
}
\label{tab:propaganda_results}
\vspace{-0.3cm}
\end{table}

\section{LLM for Detection and Explanation}
\label{sec:experiments}

\paragraph{Model.}
For developing an explanation-enhanced LLM, we adapted Llama 3.1 8B Instruct, a robust open-source model with strong multilingual capabilities \cite{dubey2024llama}. We selected the 8B variant over larger versions (70B, 45B) due to the high computational cost of fine-tuning and inference. Llama-3.1 8B has also shown strong performance in relevant multilingual tasks \cite{pavlyshenko2023analysisdisinformationfakenews, kmainasi2024llamalensspecializedmultilingualllm}.

\noindent
\paragraph{Instruction-following dataset.}
We constructed the instruction-following datasets with the aim of enhancing the model's generalizability and to guide the LLM to follow user instructions, which is a standard approach to fine-tune an LLM~\cite{zhang2023instruction}. To create versatile instructions, we prompt state-of-the-art LLMs including GPT-4o, and Claude-3.5-sonnet to generate instructions (See Appendix~\ref{apndix:prompts}). Using each LLM, we created ten diverse English instructions per language. Each instruction is uniformly distributed across dataset samples. Each sample is structured with system, user and assistant prompts, where user prompt is defined as \textit{Instruction + \{input\_text\}}, and assistant prompt is defined as \textit{Label: \{class\_label\} Explanation: \{explanation\}}.  

\noindent
\paragraph{Training.}
Due to limited computational resources, we adopted Low-rank Adaptation (LoRA)~\cite{hu2021lora} for training as a parameter-efficient fine-tuning technique. LoRA captures task-specific updates through low-rank matrices that approximate full weight updates. 

\noindent
\textbf{Parameters Setup.}  
We fine-tune the \texttt{Meta-Llama-3.1-8B-Instruct} model for 24 epochs using mixed-precision training with bfloat16 (bf16). LoRA is applied with a rank of 8, scaling factor (\(\alpha\)) of 32, and a dropout rate of 0.005. We adopt a learning rate of \(1\times10^{-4}\), following a linear warmup with a ratio of 0.05. Optimization is carried out with AdamW~\cite{loshchilov2017decoupled}, while checkpointing and evaluation are performed every 10 steps. Training employs a per-device batch size of 4, combined with gradient accumulation over 16 steps, yielding an effective batch size of 512. The maximum input sequence length is capped at 1024 tokens. All training runs are conducted on 8 NVIDIA H100 GPUs under Distributed Data Parallel (DDP)~\cite{bai2022modern}, and model selection is based on the checkpoint with the lowest validation loss.

\noindent
\textbf{Evaluation.} For the evaluation, we used a zero-shot approach and selected a random instruction from our instruction sample as a prompt, which is a common approach reported in a prior study~\cite{kmainasi2024llamalensspecializedmultilingualllm}. The temperature parameter was set to zero to ensure result reproducibility. Additionally, we implemented post-processing function to extract the labels and corresponding explanations.

\noindent
\textbf{Evaluation Metrics.} To assess classification performance, we used macro and micro F$_1$ scores. For evaluating explanations, we used BERTScore~\cite{zhang2020bertscoreevaluatingtextgeneration}, which leverages contextual embeddings. Specifically, we computed the F$_1$ score using AraBERT (v2)~\cite{antoun-etal-2020-arabert} for Arabic and BERT-base-uncased~\cite{devlin2019bert} for English.\footnote{BERTScore was chosen over BLEU and ROUGE as it captures semantics, better reflecting explanation quality.}



\section{Results and Discussion}
\label{label:results}
We compare our proposed fine-tuned Llama 3.1 8B Instruct model to baseline models: fine-tuned transformer models using AraBERT (as reported in \citet{hasanain2024can}) and BERT-base for Arabic and English, respectively. These models are commonly-used for the task~\cite{hasanain-etal-2023-araieval}. Note that BERT based models are used for label prediction only. We have also fine-tuned LLaMA without explanations to ensure a fair comparison. For Arabic, the macro-F1 score achieved is 0.780, representing a 3.0\% improvement over the fine-tuned AraBERT. For English, the results differ slightly: fine-tuned LLaMA attains a macro-F1 of 0.686, which is 0.5\% lower than that of AraBERT. 
Additionally, we compare the model's performance to two LLMs: GPT-4o and un-finetuned Llama 3.1 8B Instruct. As Table~\ref{tab:propaganda_results} shows, the performance of our fine-tuned Llama model achieves a micro F1 score that is on par or better than other models. Specifically, the model significantly outperforms the other LLMs tested.

As for its performance in explanation, in reference to the gold explanations, we observe a 25\% and 40\% improvements over the base model for English and Arabic, respectively. The fine-tuned model shows better alignment with gold explanations as demonstrated by the example in Table~\ref{tab:propaganda_results}.

  

\section{Related Work}
\label{sec:related_work}

Automatic detection of misinformation and propagandistic content has gained significant attention over the past years. Research has explored various problems, including cross-lingual propaganda analysis~\cite{barron2019proppy}, news article propaganda detection~\cite{da-san-martino-etal-2019-fine}, and misinformation and propaganda related to politics and war. Building on the seminal work of~\citet{da-san-martino-etal-2019-fine}, resources have been developed for multilingual~\cite{piskorski-etal-2023-semeval,hasanain-etal-2023-araieval} and multimodal setups~\cite{SemEval2021-6-Dimitrov,araieval:arabicnlp2024-overview}. 
Reasoning-based explanations in NLP have advanced fact-checking~\cite{russo2023benchmarking}, hate speech detection~\cite{huang-etal-2024-chatgpt}, and propaganda detection~\cite{10.1145/3613904.3642805}. While binary classifiers effectively identify propaganda, they often lack transparency, making interpretation difficult~\cite{atanasova2024generating}.~\citet{RANLP2021:propaganda:interpretable} showed that qualitative reasoning aids deception detection, while~\citet{atanasova2024generating} emphasized explanation generation for better interpretability. Yet, explicit prediction reasoning for propaganda detection remains under-explored, particularly in multilingual settings. Our work addresses the gap by developing a multilingual explanation-enhanced dataset and proposing a specialized LLM.

\section{Conclusions and Future Work}
\label{sec:conclusions}
In this study, we introduce a multilingual dataset for propaganda detection and explanation, which is the \textit{first} large dataset accompanied by explanations for the task. For Arabic, we have created a new propaganda-labeled dataset of size 13$K$ samples, consisting of tweets and news paragraphs. Using OpenAI o1, we generated explanations for this dataset, as well as for ArPro (consisting of 8$K$ instances), and for English starting from the SemEval-2023 dataset. To ensure quality, we manually evaluated the explanations and found they can serve as gold-standard references. We propose an explanation-enhanced LLM based on Llama-3.1 (8B) that matches strong baselines in performance while providing high-quality explanations.
For future work, we plan to extend it to multilabel classification and span-level propaganda  detection.



\section{Limitations}
Generating manual explanations is inherently complex. However, providing a rationale alongside the predicted label enhances trust and reliability in automated systems. Given the challenges of manual explanation creation, we relied on OpenAI's o1 -- the most capable model at the time of writing -- for generating explanations in this study. To ensure the reliability of these explanations, we conducted a manual evaluation based on four criteria: informativeness, clarity, plausibility, and faithfulness. 
The preliminary evaluation scores suggest that we can 
use them as gold explanation. For both label prediction and explanation generation, we focused on a binary classification task. However, future work should extend this to multiclass and multilabel settings. Additionally, for fine-tuning, we explored a multilingual model (Llama 3.1 8B), leaving room for further investigations into other models, including language-centric models.


\section*{Ethics and Broader Impact}
We enhanced existing datasets by incorporating explanations. To the best of our knowledge, the dataset does not include any personally identifiable information, eliminating privacy concerns. For the explanations, we provided detailed annotation guidelines. It is important to acknowledge that annotations are inherently subjective, which may introduce biases into the evaluation process. We encourage researchers and users of this dataset to critically assess these factors when developing models or conducting further studies. 


\bibliography{bibliography/main}





\appendix

\section{Annotation Guideline}
\label{sec:app_annotation_guideline}
You will be shown a news paragraph, a label assigned to it, and an explanation for the assigned label. As an annotator, your task is to carefully examine each news paragraph, label, and explanation. Then assess the quality of the explanation provided for the assigned label. Follow the steps below to ensure a thorough evaluation:

\textbf{Analyze the News Paragraph}

\begin{itemize}[noitemsep,topsep=0pt,labelsep=.5em]
    \item Read the news paragraph, sentence and/or social media post.
    \item Understand the overall message and potential implications.
\end{itemize}

\textbf{Check the Assigned Label}
\begin{itemize}[noitemsep,topsep=0pt,labelsep=.5em]
    \item Check the given label. The label is the result of annotation done by multiple human annotators.
\end{itemize}

\textbf{Evaluate the Explanation}
\begin{itemize}[noitemsep,topsep=0pt,labelsep=.5em]
    \item Read the explanation provided for why the news paragraph has been assigned its label.
    \item Assess the explanation based on the metrics below. Each metric is scored on a Likert scale from 1-5.
\end{itemize}

\subsection*{Metrics}

\paragraph{Informativeness}
Measures the extent to which the explanation provides relevant and meaningful information for understanding the reasoning behind the label. A highly informative explanation offers detailed insights that directly contribute to the justification, while a low-informative explanation may be vague, incomplete, or lacking key details.

As an annotator, you are judging if the explanation is providing enough information to explain the label assigned.

\begin{itemize}[noitemsep,topsep=0pt,labelsep=.5em]
    \item 1 = Not informative: The explanation lacks relevant details and does not help understand why the news paragraph is labeled as such.
    \item 2 = Slightly informative: The explanation provides minimal information, but key details are missing or unclear.
    \item 3 = Moderately informative: The explanation contains some useful details but lacks depth or supporting reasoning.
    \item 4 = Informative: The explanation is well-detailed, providing a clear and meaningful justification for the label.
    \item 5 = Very informative: The explanation is thorough, insightful, and fully justifies the label with strong supporting details.
\end{itemize}

\paragraph{Clarity}
Assesses how clearly the explanation conveys its meaning. A clear explanation is well-structured, concise, and easy to understand without requiring additional effort. It should be free from ambiguity, overly complex language, or poor phrasing that might hinder comprehension.

As an annotator, you are judging the language and the structure of the explanation. Spelling mistakes, awkward use of language, and wrong translation will affect this metric negatively.

\begin{itemize}[noitemsep,topsep=0pt,labelsep=.5em]
    \item 1 = Very unclear: The explanation is confusing, vague, or difficult to understand.
    \item 2 = Somewhat unclear: The explanation has some clarity but includes ambiguous or poorly structured statements.
    \item 3 = Neutral: The explanation is somewhat clear but may require effort to fully grasp.
    \item 4 = Clear: The explanation is well-structured and easy to understand with minimal ambiguity.
    \item 5 = Very clear: The explanation is highly readable, precise, and effortlessly understandable.
\end{itemize}

\paragraph{Plausibility}
Refers to the extent to which an explanation logically supports the assigned label and appears reasonable given the news paragraph's content. A plausible explanation should be coherent, factually consistent, and align with the expected reasoning behind the label. While it does not require absolute correctness, it should not contain obvious contradictions or illogical claims.

As an annotator, you are judging if the explanation actually supports the label assigned to it. For example, if a text is labeled as ``Not Propaganda,'' the explanation given should be for that label.

\begin{itemize}[noitemsep,topsep=0pt,labelsep=.5em]
    \item 1 = Not plausible at all: The explanation does not align with the label and seems completely incorrect.
    \item 2 = Weakly plausible: The explanation has some relevance but lacks strong justification or contains logical inconsistencies.
    \item 3 = Moderately plausible: The explanation somewhat supports the label but may be incomplete or partially flawed.
    \item 4 = Plausible: The explanation logically supports the label and is mostly reasonable.
    \item 5 = Highly plausible: The explanation is fully aligned with the label and presents a strong, logical justification.
\end{itemize}

\paragraph{Faithfulness}
Measures how accurately an explanation reflects the reasoning behind the assigned label. A faithful explanation correctly represents the key factors and logical steps that justify the label, without adding misleading or unrelated details. High faithfulness means the explanation stays true to the actual reasoning used for classification, ensuring reliability and consistency.

As an annotator, you are judging how well the explanation reflects the logic behind the label. For example, if the explanation claims an implication of the text, it should also present the logical reasoning behind it.

\begin{itemize}[noitemsep,topsep=0pt,labelsep=.5em]
    \item 1 = Not faithful at all: The explanation is completely unrelated to the given label and does not reflect a valid reasoning process.
    \item 2 = Weakly faithful: Some elements of the explanation are relevant, but much of it is misleading, inconsistent, or lacks proper justification.
    \item 3 = Moderately faithful: The explanation captures parts of the reasoning but includes unrelated, unclear, or unnecessary justifications.
    \item 4 = Faithful: The explanation aligns well with the reasoning behind the label and includes relevant, logical details.
    \item 5 = Highly faithful: The explanation fully and accurately reflects the correct reasoning, without any misleading or irrelevant information.
\end{itemize}

\section{Annotation Platform}
\label{sec:app_annotation_platform}

We present the screenshot of the interface designed for the evaluation of LLM generated explanation, which consisted
of a paragraph, label, and explanation for the label, annotation guidelines, and four different evaluation metrics including informativeness, clarity, plausibility, and faithfulness. 5-point Likert scale is used for each evaluation metric and the annotator is asked to follow the annotation guideline to select an appropriate Likert scale value for each metric.

\begin{figure*}[]
    \centering
    \includegraphics[scale=0.4]{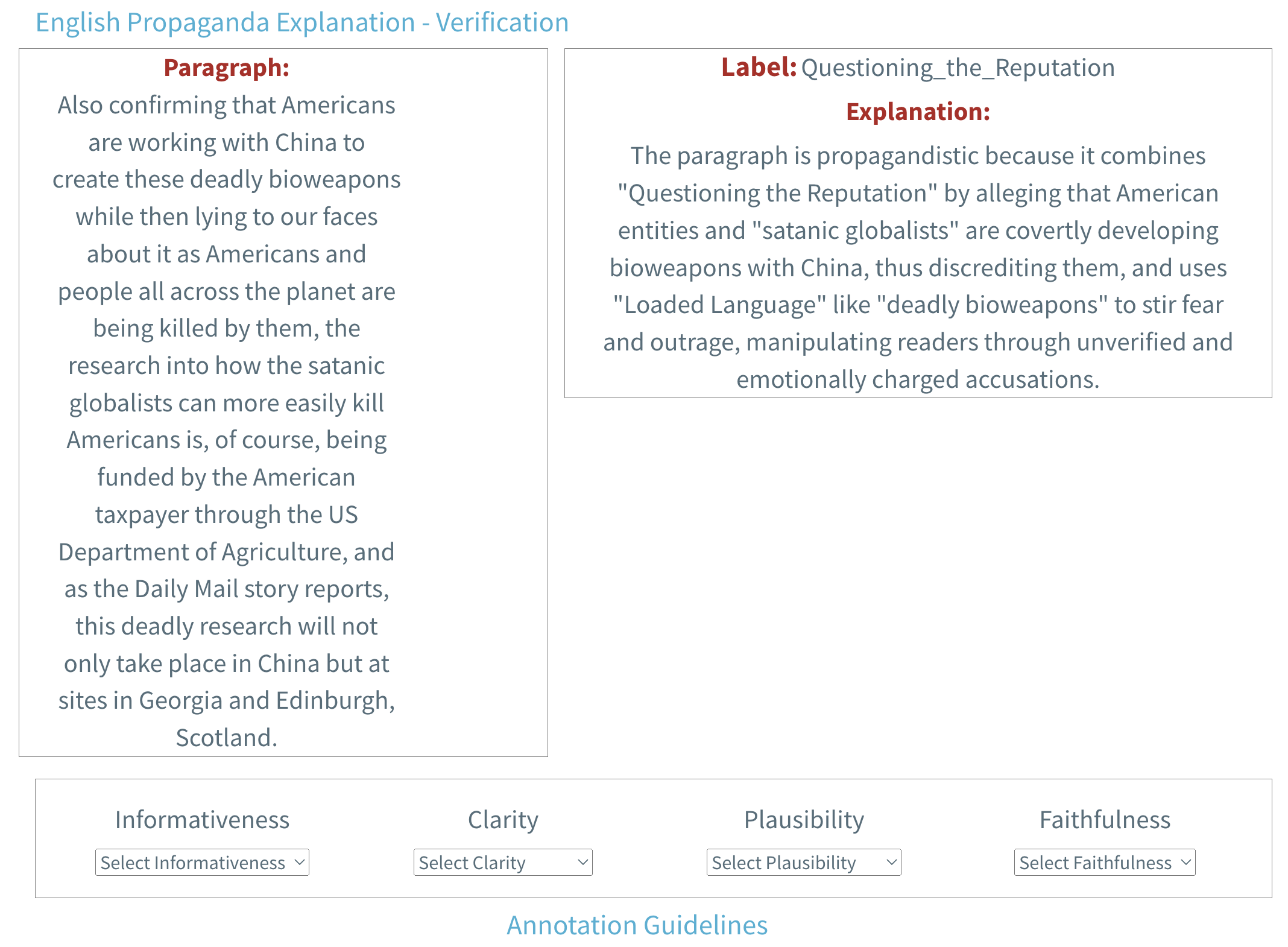}
    \caption{A screenshot of the annotation platform for the explanation evaluation of English propaganda.}
    \label{fig:hateful_meme_annotation_interface}
\end{figure*}

\section{Annotation Details}
\label{sec:app_annotation_setup}

\subsection{Annotation Setup}
We recruited annotators who are native Arabic speakers and fluent in English, all holding at least a bachelor's degree. Since they were proficient in English, they also worked on English news paragraphs. We provided annotation guidelines and necessary consultation. All annotators had prior experience with similar tasks. A total of six annotators participated in the evaluation task. In accordance with institutional requirements, each signed a Non-Disclosure Agreement (NDA). For their compensation, we hired a third-party company to manage payments at standard hourly rates based on location.

\subsection{Annotation Agreement}
\label{sec:app_annotation_agreement}
To assess the consistency of human ratings, we also computed inter-annotator agreement for each evaluation metric -- informativeness, clarity, plausibility, and faithfulness -- based on 5-point Likert scale annotations. We adopted the $r^*_{wg(j)}$ index \cite{james1984estimating}, a widely used measure for inter-annotator agreement on ordinal scales, which compares observed variance in ratings to the maximum possible variance under complete disagreement. For each item, the agreement score is computed as:
$$
r^*_{wg(j)} = 1 - \frac{S_X^2}{\sigma^2_{\text{mv}}},
$$
where $S_X^2$ is the observed variance across annotators and $\sigma^2_{\text{mv}}$ is the maximum variance possible given the scale (computed as $\sigma^2_{\text{mv}} = 0.5(X_U^2 + X_L^2) - [0.5(X_U + X_L)]^2$, with $X_U = 5$ and $X_L = 1$ for a 5-point scale). This method allows us to capture the degree of consensus among annotators while accounting for the bounded nature of Likert ratings. We report the average $r^*_{wg(j)}$ per metric. 
In Figure \ref{tab:likert_score_annotation_agr}, we report the agreement scores for both datasets. The average agreement scores for Arabic and English are above 0.89 and 0.94, respectively, for all metrics. These values indicate a strong agreement~\cite{o2017overview}.


\section{Prompts}
\label{apndix:prompts}
To generate instructions for the instruction-following dataset, we prompt the LLMs using the following prompt: \textit{We are creating an English instruction-following dataset for an [language] dataset covering the task of propaganda detection with explanation. The user defined the task as follows: Detecting propaganda in a piece of text and explaining why this piece of text is propagandistic. Propaganda can be defined as a form of communication aimed at influencing people’s opinions or actions toward a specific goal, using well-defined rhetorical and psychological techniques. For that task, the labels include: ['non-propagandistic', 'propagandistic']. Write 10 very diverse and concise English instructions making sure the labels provided above are part of the instruction. Only return the instructions without additional text.}

\section{Data Release}
\label{apndix:release}
Our proposed dataset PropXplain will be released under the CC BY-NC-SA 4.0 -- Creative Commons Attribution 4.0 International License: \url{https://creativecommons.org/licenses/by-nc-sa/4.0/}.

\section{Potential Applications}
\label{apndix:applications}

LLMs capable of detecting propaganda with explanations have several real-world applications. They can enhance social media moderation by identifying manipulative content, support fact-checkers with transparent justifications, and serve as educational tools for improving media literacy. Additionally, such models can aid NGOs and government agencies in monitoring disinformation campaigns, while offering tools to understand bias in online content. By providing interpretable outputs, these systems foster trust, accountability, and informed decision-making in digital environments.

\end{document}